\documentclass[11pt]{article}
\usepackage{coling2020}
\usepackage{times}
\usepackage{url}
\usepackage{latexsym}

\usepackage{microtype}
\usepackage{graphicx}
\usepackage{amsfonts}
\usepackage{amsmath}
\usepackage{amssymb}
\usepackage{algorithm}
\usepackage{algpseudocode}
\colingfinalcopy 

\title{Pre-trained Language Model Based Active Learning for Sentence Matching}

\author{
	Guirong Bai$^{1,2}$,
	Shizhu He$^{1,2}$,
    Kang Liu$^{1,2}$,
	Jun Zhao$^{1,2}$,
Zaiqing Nie$^3$\\
	$^1$ National Laboratory of Pattern Recognition, Institute of Automation, \\
	Chinese Academy of Sciences \\
	$^2$ School of Artificial Intelligence, University of Chinese Academy of Sciences \\
	$^3$ Alibaba AI Labs \\
	\{guirong.bai, shizhu.he, kliu, jzhao\}@nlpr.ia.ac.cn \\
	zaiqing.nzq@alibaba-inc.com
}

\date{}

\begin{document}
\maketitle
\begin{abstract}
Active learning is able to significantly reduce the annotation cost for data-driven techniques. However, previous active learning approaches for natural language processing mainly depend on the entropy-based uncertainty criterion, and ignore the characteristics of natural language. In this paper, we propose a pre-trained language model based active learning approach for sentence matching. Differing from previous active learning, it can provide linguistic criteria to measure instances and help select more efficient instances for annotation. Experiments demonstrate our approach can achieve greater accuracy with fewer labeled training instances.
\end{abstract}

\section{Introduction}

Sentence matching is a fundamental technology in natural language processing. Over the past few years, deep learning as a data-driven technique has yielded state-of-the-art results on sentence matching \cite{wang2017bilateral,chen2016enhanced,gong2017natural,yang2016anmm,parikh2016decomposable,gong2017natural,kim2019semantic}. However, this data-driven technique typically requires large amounts of manual annotation and brings much cost. If large labeled data can't be obtained, the advantages of deep learning will significantly diminish.

To alleviate this problem, active learning is proposed to achieve better performance with fewer labeled training instances \cite{settles2009active}. Instead of randomly selecting instances, active learning can measure the whole candidate instances according to some criteria, and then select more efficient instances for annotation \cite{zhang2017active,shen2017deep,erdmann-etal-2019-practical,kasai-etal-2019-low,xu-etal-2018-using}. However, previous active learning approaches in natural language processing mainly depend on the entropy-based uncertainty criterion \cite{settles2009active}, and ignore the characteristics of natural language. To be more specific, if we ignore the linguistic similarity, we may select redundant instances and waste many annotation resources. Thus, how to devise linguistic criteria to measure candidate instances is an important challenge.

\begin{figure}[htb]
\centering
\includegraphics[scale=0.50]{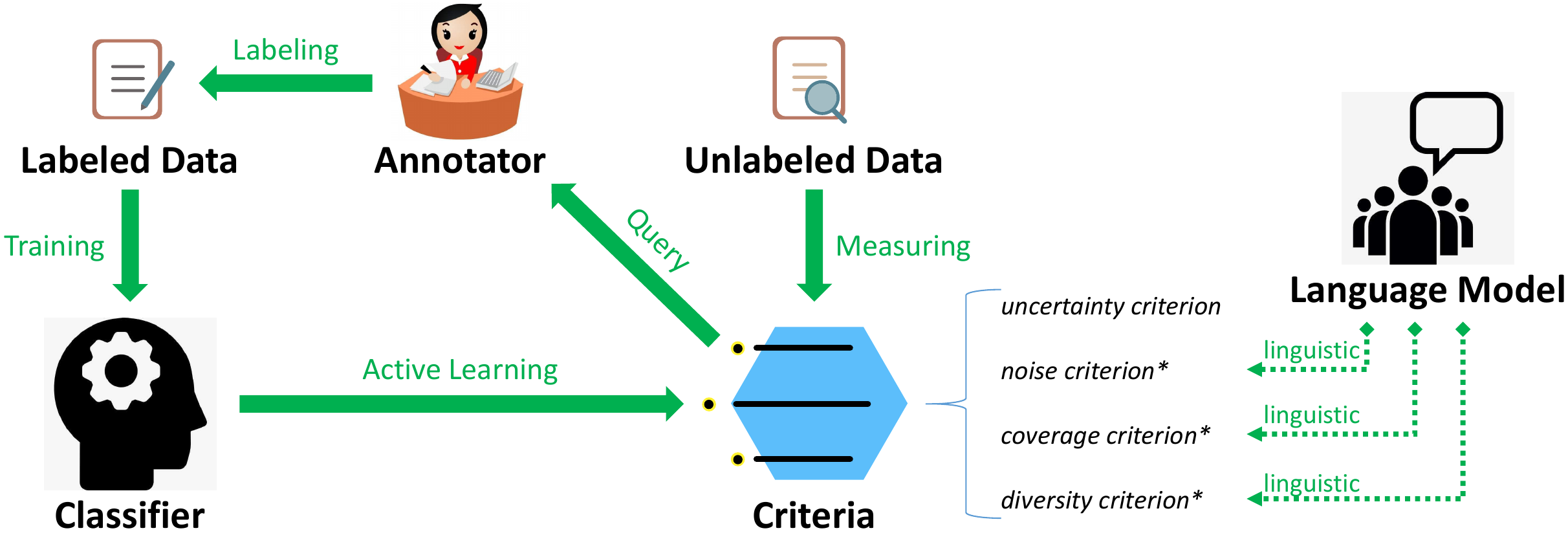}
\caption{Pipeline of our pre-trained language model based active learning. Noise, coverage and diversity are proposed linguistic criteria from a pre-trained language model.}
\label{fig:introduction}
\end{figure}

Recently, pre-trained language models \cite{peters2018deep,radford2018improving,devlin2018bert,yang2019xlnet} have been shown to be powerful for learning language representation. Accordingly, pre-trained language models may provide a reliable way to help capture language characteristics. In this paper, we devise linguistic criteria from a pre-trained language model to capture language characteristics, and then utilize these extra linguistic criteria (noise, coverage and diversity) to enhance active learning. It is shown in Figure \ref{fig:introduction}. Experiments on both English and Chinese sentence matching datasets demonstrate the pre-trained language model can enhance active learning.

\section{Methodology}

In a general active learning scenario, there is a small set of labeled training data $P$ and a large pool of available unlabeled data $Q$. Active learning is to select instances in $Q$ according to some criteria, and then label them and add them into $P$, so as to maximize classifier $M$ performance and minimize annotation cost. More details of preliminaries about sentence matching and active learning are in the Appendix.

\subsection{Pre-trained Language Model}

We choose the widely used language model BERT \cite{devlin2018bert} as the pre-trained language model. From BERT, we can obtain two kinds of information to provide linguistic criteria. One is the cross entropy loss $s_{a_i}$ of reconstructing of the $i$-th word $a_i$ in sentence $A$ (the same with another $B$) by masking only $a_i$ and predicting $a_i$ again. The other is word embeddings (contextual representations of the last layer) $\textbf{a}$=[$\textbf{e}(a_1)$,$\textbf{e}(a_2)$,\dots,$\textbf{e}(a_{l_A}$)] in the sentence, where $l_A$ is the length of sentence A.

\subsection{Criteria for Instance Selection}

\textbf{(1) Uncertainty:} The uncertainty criterion indicates classification uncertainty of an instance and is the standard criterion in active learning. Instances with high uncertainty are more helpful to optimize the classifier and thus are worthier to be selected. The uncertainty is computed as the entropy, and we can obtain uncertainty rank $rank_{uncer}(x_i)$ for the $i$-th instance in $Q$ based on the entropy. Formally,
\begin{equation}
rank_{uncer}(x_i)\propto-Ent(x_i)
\end{equation}
where $Ent(x_i)=-\sum_{k}P(y_i=k|x_i)\log P(y_i=k|x_i)$.\\
\textbf{(2) Noise:} The noise criterion indicates how much potential noise there is in an instance. Intuitively, instances with noise may degrade the labeled data $P$, and we want to select noiseless instances. Noisy instances usually have rare expression with low generating probability. Thus, noisy tokens may be hard to be reconstructed with context by the pre-trained language model. Based on this assumption, noise criterion is formulated about losses of reconstructing masked tokens:
\begin{equation}
rank_{noise}(x_i)\propto -P(A)-P(B)
\end{equation}
where $P(A)=P(a_1a_2 \dots a_{l_A})\propto\frac{l_A}{\sum_{i \in l_A}s_{a_i}}$. $P(B)$ is similar. $rank_{noise}(x_i)$ denotes noise rank of the $i$-th instance in $Q$, $s_{a_i}$/$s_{b_i}$ is the reconstruction loss of the $i$-th word $a_i$/$b_i$ in sentence $A$/$B$ from the pre-trained language model.\\
\textbf{(3) Coverage:} The coverage criterion indicates whether the language expression of the current instance can enrich representation learning. On the one hand, some tokens like stop words are meaningless and easy to model (high coverage). On the other hand, the classifier needs fresh instances (low coverage) to enrich representation learning. These fresh instances like relatively low-frequency professional expressions usually have lower generating probabilities than common ones. Thus, we can employ reconstruction losses to capture the low coverage ones as follows:
\begin{align}
&rank_{cover}(x_i)\propto -\frac{\sum_{j \in l_A}c_{a_j}s_{a_j}}{\sum_{j \in l_A}c_{a_j}}-\frac{\sum_{j \in l_B}c_{b_j}s_{b_j}}{\sum_{j \in l_B}c_{b_j}}\\
&c_{a_j}=\left\{
             \begin{array}{ll}
             {0} &if \ s_{a_j} > \beta \\
             {1} &others
             \end{array}
             \right.,c_{b_j}=\left\{
                                     \begin{array}{ll}
                                     {0} &if \ s_{b_j} > \beta \\
                                     {1} &others
                                     \end{array}
                                     \right.
\end{align}
where $\beta$ denotes a hyperparameter to distinguish noise and is set as 10.0.\\
\textbf{(4) Diversity:} The diversity criterion indicates the diversity of instances. Redundant instances are inefficient and waste annotation resources. In contrast, diverse ones can help learn more various language expressions and matching patterns.

First, we use a vector $\textbf{v}_i$ for instance representation of a sentence pair instance $x_i$. To model the difference between two sentences, we employ the subtraction of word embeddings between ``Delete Sequence" $L_D$ and ``Insert Sequence" $L_I$ from Levenshtein Distance (when we transform sentence $A$ to sentence $B$ by deleting and inserting tokens, these tokens are added into $L_D$ and $L_I$ respectively). It is illustrated in the Appendix. Besides, the word embeddings in the subtraction are weighted by reconstruction losses. Intuitively, meaningless tokens such as preposition should have less weight, and they are usually easier to predict with lower reconstruction losses. Formally,
\begin{align}
\textbf{v}_i=\sum_{j \in L_I}w_{b_j}\textbf{e}(b_j)-\sum_{j \in L_D}w_{a_j}\textbf{e}(a_j)\\
w_{a_j}=\frac{s_{a_j}}{\sum_{k \in l_A}s_{a_k}},w_{b_j}=\frac{s_{b_j}}{\sum_{k \in l_B}s_{b_k}}
\end{align}
where $s_{a_i}$/$s_{b_j}$ is the reconstruction loss of the $i$/$j$-th word of sentence $A$/$B$. $\textbf{e}(a_j)$/$\textbf{e}(b_j)$ denotes word embdeddings. $w_{a_i}$/$w_{b_j}$ denotes the weight for tokens.

With instance representation, we want to select diverse ones that are representative and different from each other. Specifically, we employ k-means clustering algorithm for diversity rank as follows:
\begin{align}
rank_{diver}(x_i)&=\left\{
             \begin{array}{ll}
             {0} &if \ \textbf{v}_i\circ \textbf{v}_i \in O_{diver} \\
             {n} &others
             \end{array}
             \right.
\end{align}
where $O_{diver}$ are the centers of $n$ clusters of $\{\textbf{v}_i\circ \textbf{v}_i\}$. $\circ$ denotes multiplication on element.

\subsection{Instance Selection}

In practice, according to different effectiveness of criteria, we combine ranks of criteria and select the top $n$ candidate instances in unlabeled data $Q$. Specifically, we sequentially use $rank_{uncer}$, $rank_{diver}$, $rank_{cover}$, $rank_{noise}$ to select top $8n$, $4n$, $2n$, $n$ candidate instances, and add the final $n$ instances into labeled data $P$ for training at every round.

\section{Experiments}

\subsection{Settings and Comparisons}

We conduct experiments on Both English and Chinese datasets, including \textbf{SNLI} \cite{bowman2015large}, \textbf{MultiNLI} \cite{williams2017broad}, \textbf{Quora} \cite{iyer2017first}, \textbf{LCQMC} \cite{liu2018lcqmc}, \textbf{BQ} \cite{chen2018bq}. The number of instances to select at every round is $n=100$. We choose \cite{devlin2018bert} as classifier $M$ and perform 25 rounds of active learning. There is a held-out test set for evaluation after all rounds. We compare the following active learning approaches:\\
\textbf{(1)Random sampling (Random)} randomly selects instances for annotation and training at each round.\\
\textbf{(2)Uncertainty sampling (Entropy)} is the standard entropy criterion \cite{tong2001support,Zhu2008Active}.\\
\textbf{(3)Expected Gradient Length (EGL)} aims to select instances expected to result in the greatest change to the gradients of tokens. \cite{settles2008analysis,zhang2017active}.\\
\textbf{(4)Pre-trained language model (LM)} is our proposed active learning approach.

\subsection{Results}

\begin{table}[htb]\footnotesize
\centering
\begin{tabular}{c|ccccc}
\hline
&SNLI&MultiNLI&Quora&LCQMC&BQ \\
\hline
Random & 77.90 & 67.83 & 79.01 & 82.04 & 71.44\\
\hline
Entropy & 79.80 & 70.27 & 80.21 & 83.25 & 73.60\\
\hline
EGL & 77.86 & 66.80 & 77.91 & 80.35 & 71.59\\
\hline
LM & \textbf{80.99} & \textbf{71.79} & \textbf{81.79} & \textbf{84.29} & \textbf{74.73}\\
\hline
\hline
& Ent & E+Cov & E+Noi & E+Div & E+All \\
\hline
Ablation&79.80 & 80.99 & 81.11 & \textbf{81.45} & 80.99 \\
\hline
\end{tabular}
\caption{The upper part lists accuracy of different approaches on five datasets. The low part lists accuracy of combining different linguistic criterion with uncertainty on SNLI dataset for ablation.}
\label{tab:total accuracy}
\end{table}

\begin{figure}[htb]
\centering
\includegraphics[scale=0.39]{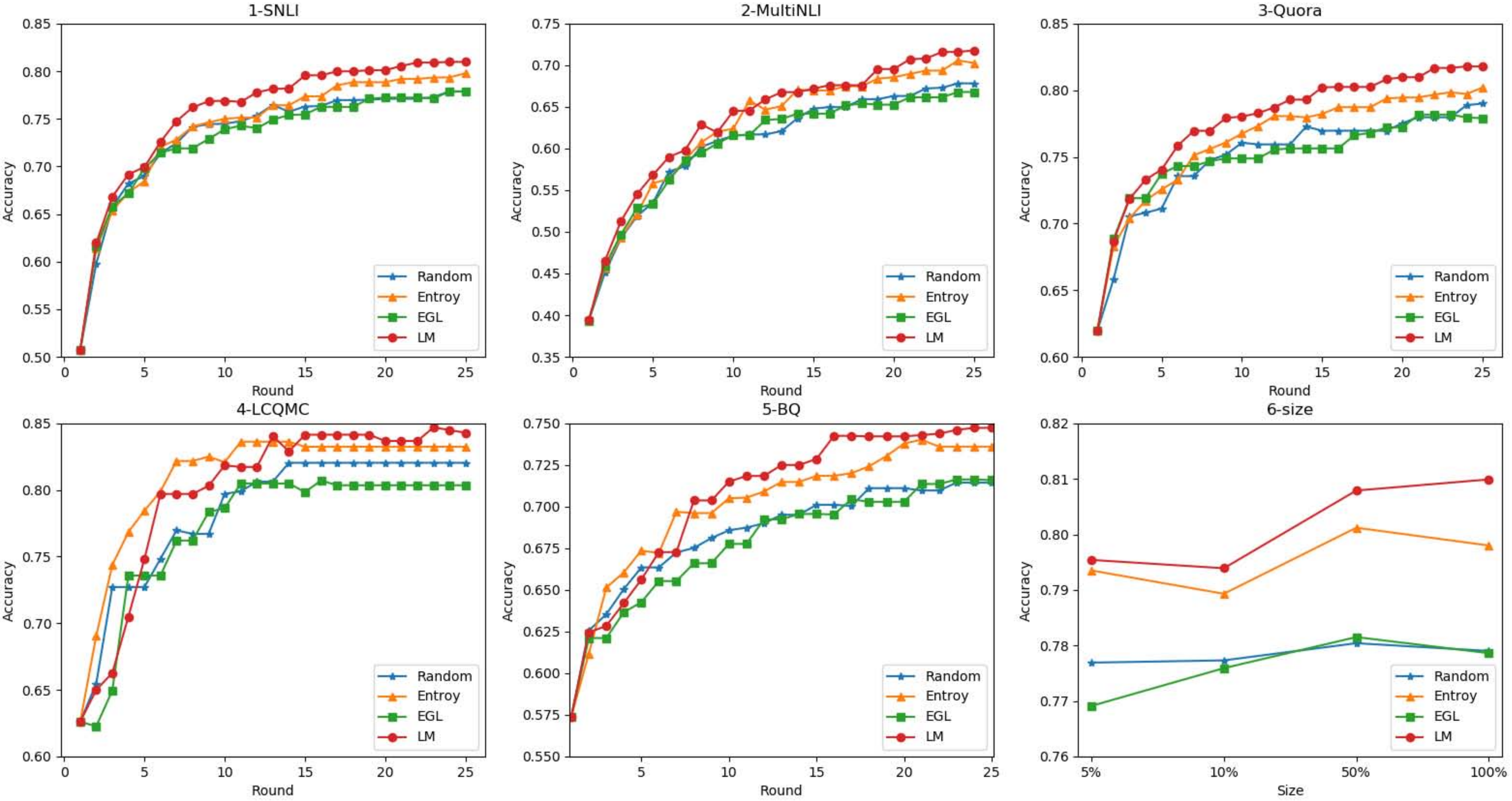}
\caption{The figures 1-5 are learning curves of comparisons on the five datasets. The 6-th figure illustrates learning curves on four SNLI subsets to show the relation between data size and accuracy.}
\label{fig:total curves}
\end{figure}
Table \ref{tab:total accuracy} and Figure \ref{fig:total curves} (1-5) report accuracy and learning curves of each approach on the five datasets. Overall, our approach obtains better performance on both English and Chinese datasets. We can know that extra linguistic criteria are effective, demonstrating that a pre-trained language model can substantially capture language characteristics and provide more efficient instances for training. Besides, active learning approaches always obtain better performance than random sampling. It demonstrates that the amount of labeled data for sentence matching can be substantially reduced by active learning. And EGL performs worse than the standard approach active learning, maybe gradient based active learning is not suitable for sentence matching. In fact, sentence matching needs to capture the difference between sentences and gradients of a single token can't reflect the relation. Moreover, we show the relation between the size of unlabeled data and accuracy in Figure \ref{fig:total curves} (6), we can see the superiority of the pre-trained model based approach is more significant for larger data size.

\subsection{Ablation Study}

To validate the effectiveness of extra linguistic criteria, we separately combining them with standard uncertainty criterion. ``Ent'' denotes the standard uncertainty criterion, ``E+Noi/E+Cov/E+Div/E+All'' denotes combining uncertainty with noise/coverage/diversity/all criteria. Table \ref{tab:total accuracy} reports the accuracy. Curves are also illustrated in the Appendix.

We can see each combined criterion performs better than a single uncertainty criterion. It demonstrates that each linguistic criterion from a pre-trained language model helps capture language characteristics and enhances selection of instances. More ablation discussions are shown in the Appendix.

%

\section{Conclusion}

In this paper, we combine active learning with a pre-trained language model. We devise extra linguistic criteria from a pre-trained language model, which can capture language characteristics and enhance active learning. Experiments show that our proposed active learning approach obtains better performance.

\section*{Acknowledgements}

The work is supported by the National Natural Science Foundation of China under Grant Nos.61533018, U1936207, 61976211, and 61702512, and the independent research project of National Laboratory of Pattern Recognition under Grant. This research work was also supported by Youth Innovation Promotion Association CAS.
\bibliographystyle{coling}
\bibliography{coling2020}
\appendix
\newpage
\noindent\textbf{Appendix A: More Details and Discussions}\\
\textbf{Sentence Matching Task:} Given a pair of sentences as input, the goal of the task is to judge the relation between them, such as whether they express the same meaning. In formal, we have two sentences $A$=[$a_1$,$a_2$,\dots,$a_{l_A}$] and $B$=[$b_1$,$b_2$,\dots,$b_{l_B}$], where $a_i$ and $b_j$ denote the $i$-th and $j$-th word respectively in corresponding sentences, and $l_A$ and $l_B$ denote the length of corresponding sentences.

Through a shared word embedding matrix $\textbf{W}_e$ $\in$ $\mathbb{R}^{n_e \times d}$, we can obtain word embeddings of input sentences $\textbf{a}$=[$\textbf{e}(a_1)$,$\textbf{e}(a_2)$,\dots,$\textbf{e}(a_{l_A}$)] and $\textbf{b}$=[$\textbf{e}(b_1)$,$\textbf{e}(b_2)$,\dots,$\textbf{e}(b_{l_B})$], where $n_e$ denotes the vocabulary size, $d$ denotes the embedding size and $\textbf{e}(a_i)$ and $\textbf{e}(b_j)$ denote the word embedding of the $i$-th and $j$-th word respectively in corresponding sentences. And there is a sentence matching model $M$ to predict a label $\hat{y}$ based on $\textbf{a}$ and $\textbf{b}$. When testing, we choose the label with the highest probability in prediction distribution $P(y_i|\textbf{a},\textbf{b};\theta_M)$ as output, where $\theta_M$ denotes parameters of the model $M$ and $y_i$ denotes a possible label. When training, the model $M$ is optimized by minimizing cross entropy:
\begin{equation}
Loss=-P(y|\textbf{a},\textbf{b};\theta_M)\log P(y|\textbf{a},\textbf{b};\theta_M)
\end{equation}
where $y$ denotes the golden label.\\
\textbf{Standard Active Learning:} In a general active learning scenario, there exists a small set of labeled data $P$ and a large pool of available unlabeled data $Q$. $P$ is for training a classifier and can absorb new instances from $Q$. The task for the active learning is to select instances in $Q$ based on some criteria, and then label them and add them into $P$, so as to maximize classifier performance and minimize annotation cost. In the selection criteria, a measure is used to score all candidate instances in $Q$, and instances maximizing this measure are selected into $P$.

The process is illustrated in Algorithm \ref{alg:Active learning}. The instance selection process is iterative, and the process will repeat until a fixed annotation budget is reached. At every round, there are $n$ instances to be selected and labeled.
\begin{algorithm}[htb]
\caption{Active learning algorithm flow.}
\label{alg:Active learning}
\begin{algorithmic}
\Require\\
labeled data set $P$=\{$\varnothing$\}, unlabeled data set $Q$=\{$q_i$\}, the classifier $M$, criteria of instance selection $C$, the number of instances for annotation at every round $n$
\end{algorithmic}
\begin{algorithmic}
\Ensure\\
labeled data set $P$=\{$p_i$\}, the classifier $M$
\end{algorithmic}
\begin{algorithmic}[1]
\Repeat
\State Sort $Q$ based on $M$ and $C$
\State Select top $n$ instances from $Q$ to label, update $Q$
\State Add labeled $n$ instances into $P$, update $P$
\State Train and update classifier $M$ based on $P$
\Until{The annotation budget is exhausted}
\end{algorithmic}
\end{algorithm}

With the same amount of labeled data $P$, criteria for instance selection in active learning determine the classifier performance. Commonly, the criteria is mainly based on uncertainty criterion ($uncertainty$ $sampling$), in which ones near decision boundaries have priority to be selected. A general uncertainty criterion uses entropy, which is defined as follows:
\begin{equation}
Ent(x_i)=-\sum_{k}P(y_i=k|x_i)\log P(y_i=k|x_i)
\label{eq:uncertainty}
\end{equation}
where $k$ indexes all possible labels, $x_i$ denotes a candidate instance that is made up of a pair of sentences $A$ and $B$ in available unlabeled data Q.
\begin{figure}[h]
\centering
\includegraphics[scale=0.55]{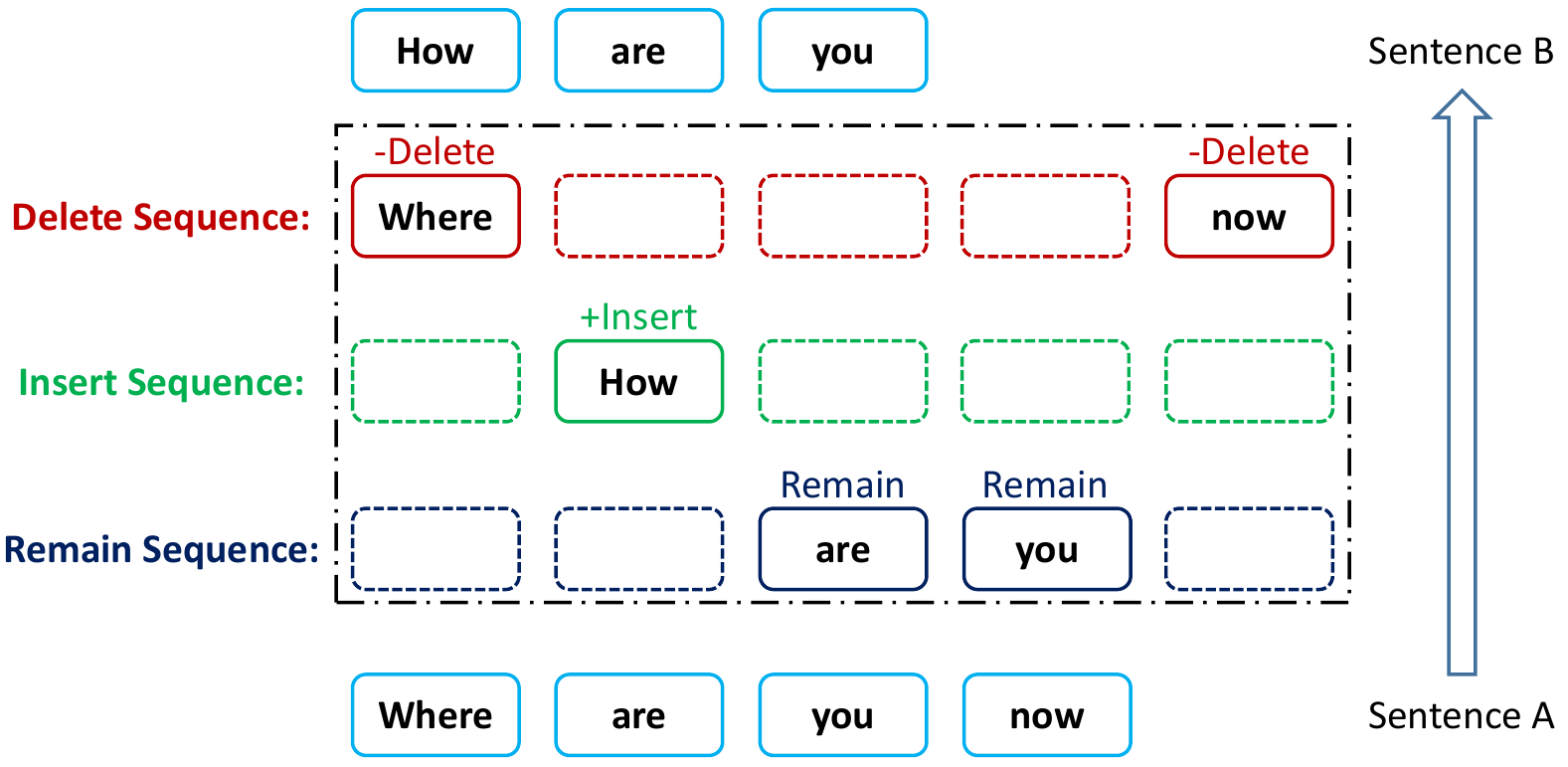}
\caption{``Delete Sequence" and ``Insert Sequence".}
\label{fig:edit}
\end{figure}

\noindent\textbf{Visualization of Delete Sequence and Insert Sequence:} To model the difference between two sentences, we employ the subtraction of word embeddings between ``Delete Sequence'' and ``Insert Sequence'' from Levenshtein Distance (when we transform sentence A to sentence B by deleting and inserting tokens, these tokens are added into ``Delete Sequence'' and ``Insert Sequence'' espectively). We illustrate it in Figure \ref{fig:edit}.\\
\textbf{Datasets:} We conduct experiments on three English datasets and two Chinese dataset. Table \ref{tab:statistics of datasets} provides statistics of these datasets.

\textbf{(1)SNLI:} an English natural language inference corpus based on image captioning.

\textbf{(2)MultiNLI:} an English natural language inference corpus with greater linguistic difficulty and diversity.

\textbf{(3)Quora:} an English question matching corpus from the online question answering forum Quora.

\textbf{(4)LCQMC:} an open-domain Chinese question matching corpus from the community question answering website Baidu Knows.

\textbf{(5)BQ:} an in-domain Chinese corpus question matching corpus from online bank custom service logs.
\begin{table}[htb]
\centering
\begin{tabular}{|c|ccc|}
\hline
& training & validation & test \\
\hline
SNLI & 549,367 & 9,842 & 9,824 \\
\hline
MultiNLI & 392,702 & 9,815 & 9,832 \\
\hline
Quora & 384,348 & 10,000 & 10,000 \\
\hline
LCQMC & 238,766 & 8,802 & 12,500 \\
\hline
BQ & 100,000 & 1,000 & 1,000 \\
\hline
\end{tabular}
\caption{Statistics of sentence matching datasets.}
\label{tab:statistics of datasets}
\end{table}
\begin{figure}[h]
\centering
\includegraphics[scale=0.5]{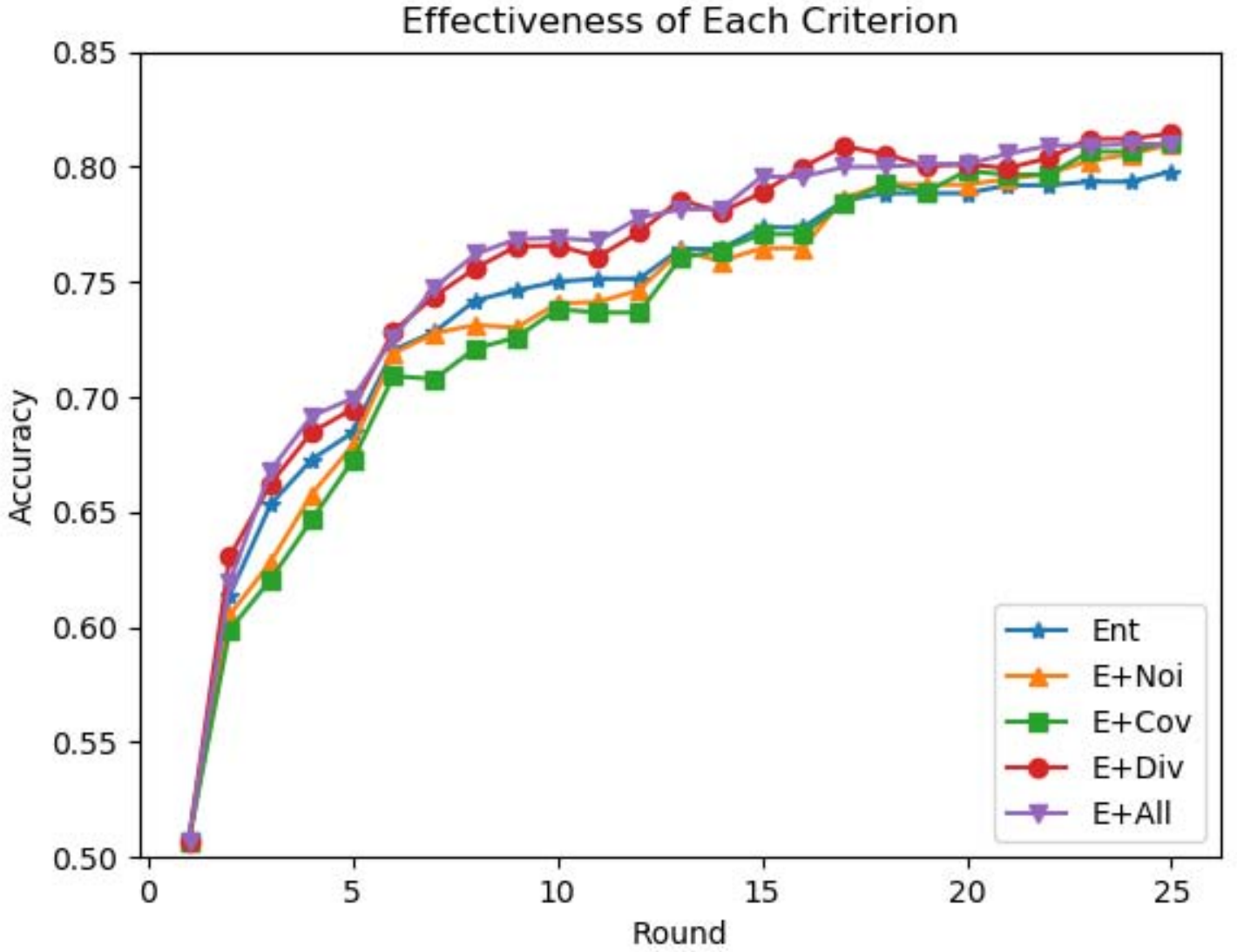}
\caption{Learning curves of combining each proposed linguistic criterion with uncertainty on SNLI dataset.}
\label{fig:criterion effectiveness}
\end{figure}
\begin{figure}[t]
\centering
\includegraphics[scale=0.5]{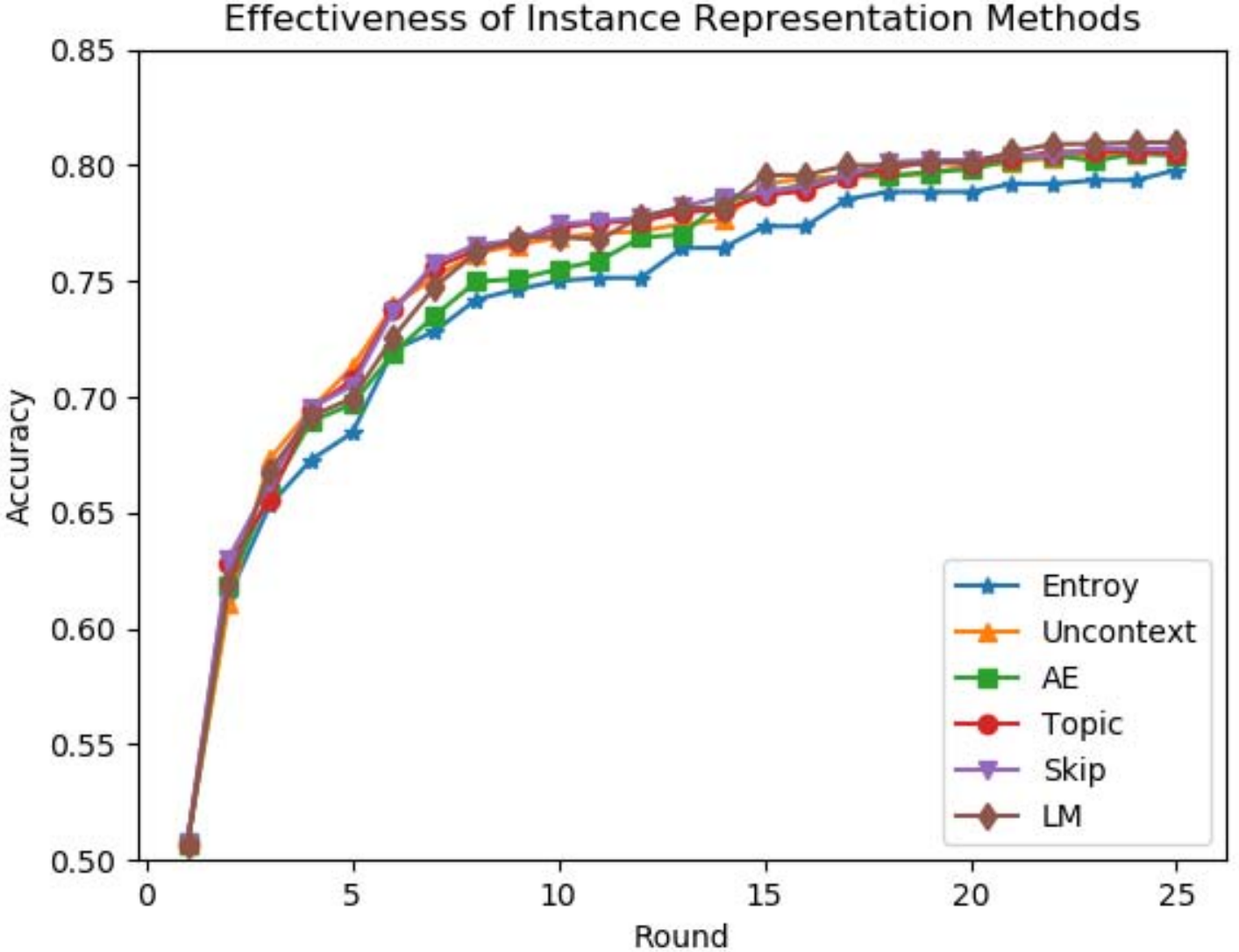}
\caption{Learning curves of different instance representation methods.}
\label{fig:vector effectiveness}
\end{figure}
\begin{figure}[t]
\centering
\includegraphics[scale=0.5]{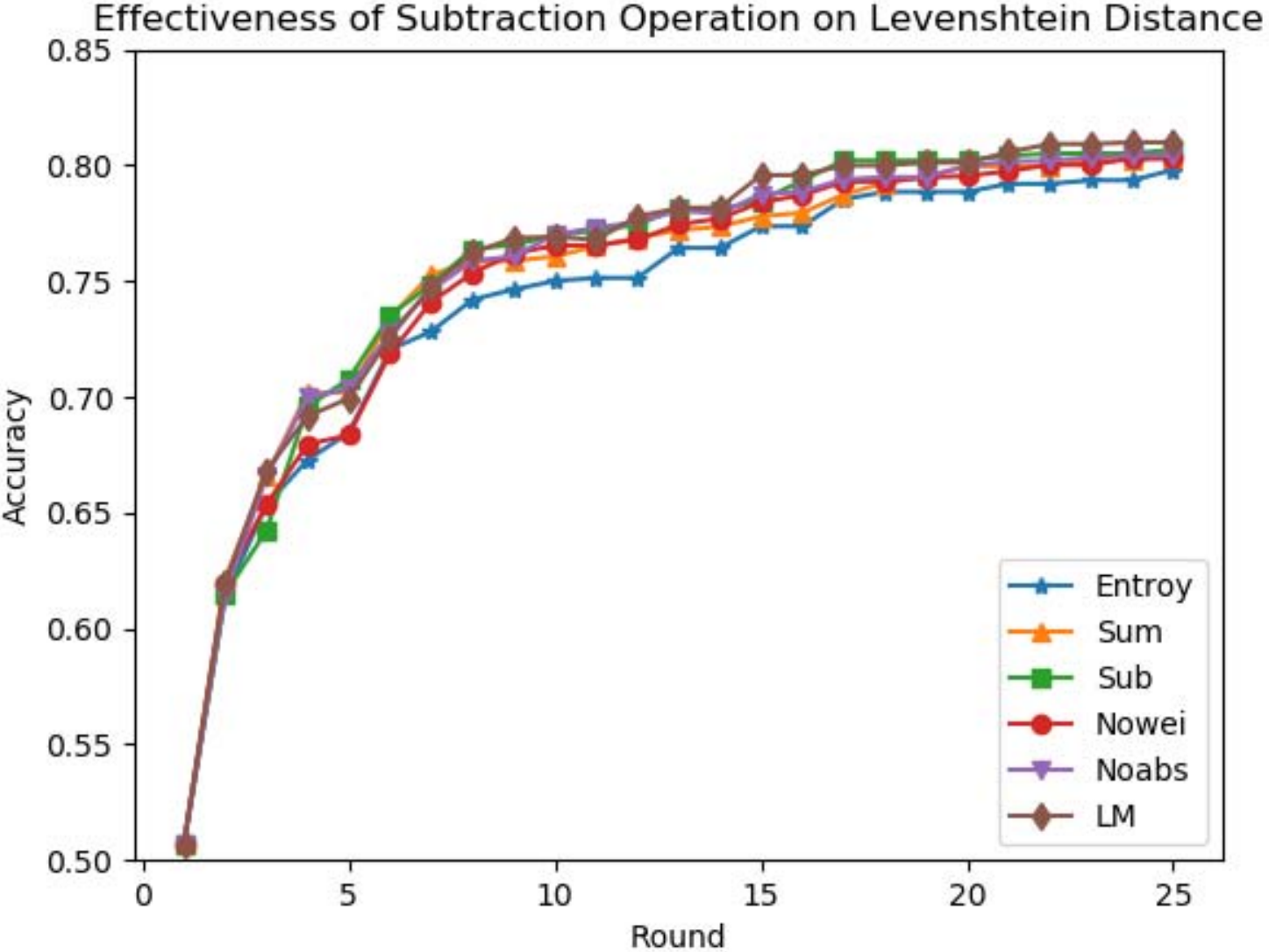}
\caption{Learning curves of subtraction operation on Levenshtein Distance.}
\label{fig:operation effectiveness}
\end{figure}

\noindent\textbf{Configuration:} The number of instances to select $n$ is 100 at every round and we perform 25 rounds of active learning, that is there are total of 2500 labeled instances for training in the end. Batch size is 16 for English and 32 for Chinese, Adam is used for optimization. We evaluated performance by calculating accuracy and learning curves on a held-out test set (classes are fairly balanced in datasets) after all rounds.\\
\textbf{Curves of Ablation Study:} Figure \ref{fig:criterion effectiveness} shows learning curves of combining each proposed linguistic criterion with uncertainty on SNLI dataset.

\noindent\textbf{Discussion:}

\textbf{(1)Effectiveness of different instance representation methods:} We validate the effectiveness of different instance representation methods in diversity criterion on SNLI dataset. We compare our method with 4 baselines: (a) using the first word embedding layer in BERT as context-dependent representations (Uncontext); (b) using the subtraction between sentence vectors from auto-encoding (AE); (c) using the subtraction between sentence vectors from topic model (Topic); (d) using the subtraction between sentence vectors from Skip-Thoughts (Skip).
\begin{table}[htb]\footnotesize
\centering
\begin{tabular}{|cccccc|}
\hline
Entroy & Uncontext & AE & Topic & Skip & LM \\
\hline
79.80 & 80.63 & 80.42 & 80.54 & 80.71 &\textbf{80.99} \\
\hline
\end{tabular}
\caption{Accuracy of different instance representation methods.}
\label{tab:vector effectiveness}
\end{table}

Table \ref{tab:vector effectiveness} and Figure \ref{fig:vector effectiveness} report accuracy and learning curves respectively. We can see contextual representations are better than context-dependent representations. In intuition, contextual representations are more exact especially when dealing with polysemy. Next, we find our proposed method outperforms sentence vector based methods (Topic, AE, and Skip). It is possibly because BERT used more data to learn language representations.

\textbf{(2)Effectiveness of subtraction operation on Levenshtein Distance:}
Here we validate the effectiveness of the operation that uses the subtraction of word embeddings between ``Delete Sequence" and ``Insert Sequence" in diversity criterion on SNLI dataset. We compare it with 4 baselines: (a) using the sum of word embeddings of the two sentences (Sum); (b) directly using the subtraction of word embeddings of the two sentences without ``Delete Sequence" and ``Insert Sequence" (Sub); (c) without weight for word embeddings (Nowei); (d) without absolute value operation for symmetry (Noabs).
\begin{table}[htb]\footnotesize
\centering
\begin{tabular}{|cccccc|}
\hline
Entroy & Sum & Sub & Nowei & Noabs & LM \\
\hline
79.80 & 80.35 & 80.67 & 80.29 & 80.44 & \textbf{80.99} \\
\hline
\end{tabular}
\caption{Accuracy of subtraction operation on Levenshtein Distance.}
\label{tab:operation effectiveness}
\end{table}

Table \ref{tab:operation effectiveness} and Figure \ref{fig:operation effectiveness} report accuracy and learning curves respectively. We can see subtraction operation is better than sum operation. It demonstrates that subtraction has better ability to capture the difference between two sentences, and provides better instance representation for diversity rank. We can see the results without ``Delete Sequence" and ``Insert Sequence" performs a little worse, proving its necessity. And the results without weight operation for word embeddings perform worse. We can know weight for meaningless tokens is effective. Besides, we can see the results without absolute value operation for symmetry is worse, demonstrating absolute value operation is necessary.

\end{document}